# Classification of Basmati Rice Grain Variety using Image Processing and Principal Component Analysis

Rubi Kambo[#1], Amit Yerpude[*2]

[#]*M.Tech Scholar, Department Of Computer Science and Engg., CSVTU University*
*Rungta College of Engg. & Technology, Bhilai (C.G), INDIA*

[*]*Associate Professor, Department Of Computer Science and Engg., CSVTU University*
*Rungta College of Engg. & Technology, Bhilai (C.G), INDIA*

*Abstract* — **All important decisions about the variety of rice grain end product are based on the different features of rice grain. There are various methods available for classification of basmati rice. This paper proposed a new principal component analysis based approach for classification of different variety of basmati rice. The experimental result shows the effectiveness of the proposed methodology for various samples of different variety of basmati rice.**

## I. INTRODUCTION

Rice is the seed of the monocot plants Oryza sativa (Asian rice) [1] [2]. In Asia, especially as a cereal grain, it is the most widely consumed staple food for a large part of the world's human population [3]. After corn, it is the grain with the second-highest worldwide production, according to data for 2010 "Prod STAT" by FAOSTAT (Food And Agriculture Organization Of The United Nations) Retrieved December 26, 2006).

Basmati rice means the "queen of fragrance or the perfumed one." According to the Agricultural and Processed Food Products Export Development Authority (APEDA), India is the second largest producer of rice after China, and grows over a tenth of the world's wheat. In 1993, Basmati rice attracted the highest premium because it is very-long grained rice, with an aroma of its own which enhances the flavors it's mixed with.

"Basmati" is long grain aromatic rice grown for many centuries in the specific geographical area, at the Himalayan foot hills of Indian sub-continent, blessed with characteristics extra- long slender grains that elongate at least twice of their original size with a characteristics soft and fluffy texture upon cooking, delicious taste, superior aroma and distinct flavor, Basmati rice is unique among other aromatic long grain rice varieties. Having an average length of about 7.00mm [4] [5].

The main varieties of Basmati rice as notified under the seeds Act, 1966 are Basmati 386 , Basmati 217 , Ranbir Basmati , Karnal Local/ Taraori Basmati, Basmati 370, Type-3 (Dehradooni Basmati), Pusa Basmati-1, Pusa Basmati 1121, Punjab Basmati-1, Haryana Basmati- 1, Kasturi and Mahi Sugandha.

An excellent quality Indian Basmati Rice that includes Kohinoor Basmati Rice, India Gate Basmati Rice, Daawat Basmati Rice etc. Offered rice is hygienically processed under the supervision of highly experienced professionals at vendors' end. Owing to delectable taste, high nutritional value and rich aroma, offered rice is highly appreciated by the customer. To preserve its natural taste and quality, we utilize finest quality packaging material to pack the entire this rice.

India gate Basmati brand has different variety like Classic, Rozana and mini dubar. India gate classic is an exotic class of basmati. It embodies all the attributes of true basmati rice grain. Its smooth, pearl white grain , extra fine and extra long.

High-dimensional datasets present many mathematical challenges as well as some opportunities, and are bound to give rise to new theoretical developments. For understanding the underlying phenomena of interest, one of the problems with high-dimensional datasets is that, in many cases, not all the measured variables are important.

Dimensionality Reduction is about transforming data of very high dimensionality into data of much lower dimensionality such that each of the lower dimensions convey much more information and can be divided into feature selection and feature extraction.

## II. PRINCIPAL COMPONENT ANALYSIS

Impact of PCA is affecting the research work in now a day in the various fields like application of Image Processing, pattern recognition, Neural network and etc. PCA is a powerful and widely used linear technique in statistics, signal processing, image processing, and elsewhere. In statistics, PCA is a method for simplifying a multidimensional dataset to lower dimensions for analysis, visualization or data compression [6].

The Principal Component Analysis (PCA) is one of the most successful techniques that have been used in image recognition and compression. The purpose of PCA is to reduce the large dimensionality of the data space (i.e. observed variables) to the smaller intrinsic dimensionality of feature space (i.e. Independent variables), which are needed to describe the data economically. This is the case when there is a strong correlation between observed variables.

Principal component analysis (PCA) is a classical statistical method. This linear transform has been widely used in data analysis and compression. Principal component analysis is based on the statistical representation of a random variable [6].PCA involves the calculation of the Eigen value decomposition of a data covariance matrix or singular value decomposition of a data matrix, usually after mean centring the data for each attribute. It is the simplest of the true eigenvector based multivariate analyses. Often, its operation can be thought of as revealing the internal structure of the data in a way which best explains the variance in the data.

There are some steps for implementing Principal Component Analysis. They are:-

**Step-1**-Take an original data set and calculate mean of the data set taking as column vectors, each of which has M rows. Place the column vectors into a single matrix X of dimensions M × N.

**Step-2**-Subtract off the mean for each dimension. Find the empirical mean along each dimension m = 1... M of each column. Place the calculated mean values into an empirical mean vector u of dimensions M × 1.

$$u[m] = (1/N) \sum_{n=1}^{N} X[m, n]$$

Mean subtraction is an integral part of the solution towards finding a principal component basis that minimizes the mean square error of approximating the data. There are two steps:

1. Subtract the empirical mean vector u from each column of the data matrix X.
2. Store mean-subtracted data in the M*N matrix B.

**B=X-uh**

[Where h is a 1 x N row vector of all 1's: h[n] =1 for n=1…N]

**Step-3**-Calculate the covariance matrix Find the M × M empirical covariance matrix C from the outer product of matrix B with itself: -

**C=E [B × B*] =E [B.B*] = (1/N) ∑ B.B***

Where E is the expected value operator, × is the outer product operator, and * is the conjugate transpose operator. Note that if B consists entirely of real numbers, which is the case in many applications, the "conjugate transpose" is the same as the regular transpose.

**Step-4**-Calculate eigenvector and Eigen value of the covariance matrix compute the matrix V of eigenvectors which diagonalizes the covariance matrix

**C: V-1CV=D**

Where D is the diagonal matrix of Eigen values of C . This step will typically involve the use of a computer-based algorithm for computing eigenvectors and eigenvalues.

**Step-5**-Extract diagonal of matrix as vector:

Matrix D will take the form of an M × M diagonal matrix, where

**D [p, q] =λ m**

for p=q=m is the mth Eigen value of the covariance matrix C, and

**D [p, q] =0**

For p≠q Matrix V, also of dimension M × M, contains M column vectors, each of length M, which represent the M eigenvectors of the covariance matrix C.

**Step-6**-Sorting in variance in decreasing order Sort the column of the Eigen vector matrix V and Eigen value matrix D in order of decreasing Eigen value .

**Step-7**-Choosing components and forming a feature vector here is where the notion of data compression and reduced dimensionality comes into it.

If we look at the eigenvectors and Eigen values from the previous section, we will observe that the Eigen values are quite different values.

In fact, it turns out that the eigenvector with the highest Eigen value is the principle component of the data set. What needs to be done now is you need to form a feature vector.

Taking the eigenvectors that we want to keep from the list of eigenvectors, and forming a matrix with these eigenvectors in the columns construct this: -

**Feature Vector = (eig1 eig2 eig3……….eig n)**

**Step-8**-Deriving the new data set, is the final step in PCA and it is also the easiest. Once we have chosen the components (eigenvectors) that we wish to keep in our data and formed a feature vector, we simply take the transpose of the vector and multiply it on the left of the original data set, transposed.

**Final Data=Row Feature Vector*Row Data**

Adjust where Row Feature Vector is the matrix with the eigenvectors in the columns transposed so that the eigenvectors are now in the rows, with the most significant eigenvectors at the top, and Row Data Adjust is the mean-adjusted data transposed, i.e. the data items are in each column, with each row holding a separate dimension [7].

## III. MATERIALS AND METHODOLOGY

From Supermarket store, basmati rice grain samples were collected and for rice granules of different size Samsung mobile camera is used to acquire and record the images .In this rice grain is placed on the black sheet of paper [8] .To collect image data, the camera is placed at fixed location and mounted on stand, to provide vertical movement and the distance between the lens and the sample table is 14 cm .The Black background is used and light intensity on sample table is uniform to improve the data collection under controlled environment. All the grains in the sample image were arranged in arbitrary direction and position. Images were captured and stored in JPG format automatically .Through data cables these images has been transferred and then stored in hard disk and then different parameters of rice were extracted from the image for further analysis.

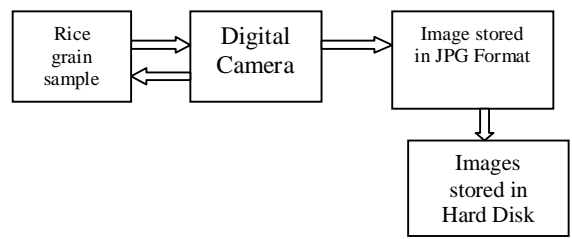

Fig. 1 Basic Building block for image capturing [8] [9]

IV. IMAGE ANALYSIS

A. *Image Acquisition*

Image acquisition in image processing can be broadly defined as the action of retrieving an image from some source. Image acquisition in image processing is always the first step in the workflow sequence because, without an image, no processing is possible. Acquisition of image can be done under uniform lighting by Samsung mobile Digital camera [10] [9] [11].

B. *Image Pre-processing and Smoothing.*

The aim of pre-processing is an improvement of image data that suppresses unwanted distortion or enhances some image features for further processing. For human viewing, Image Enhancement improves the quality and clarity of images. Removing noise and blur, rising contrast and enlightening details from images are example of enhancement operation. Noise tends to attack images when picture are taken in low light setting.

While capturing the image, sometime it has been distorted and hence image is to be enhanced by applying special median filtering to the image to remove noise [12].Filtering types ,noise reduction techniques such as Averaging, Gaussian filters are used and causes image smoothing .In this paper ,Median filter is used for smoothing because it protect the edges of the image during noise removal and is mostly used in digital imaging and effective with salt and pepper noise and speckle noise .The noise in the input gray color image is detached using median filter [10].

C. I*mage Segmentation*

After image enhancement, the next process in image processing is the image segmentation and the very first step in image analysis is image segmentation where the image is subdivided into different parts or object. Basically the image is subdivided until we segregate the interested object from their background. Generally there are two approaches for segmentation algorithms. one is based on the discontinuity of gray level values and the other is based on the similarity of gray level values and for this different approaches like thresholding, region growing ,region splitting and merging can be used [12] [10] [9] [13]. Image segmentation is typically used to locate objects and boundaries in images.

Segmentation can also be done using edge detection. Edge detector detect the discontinuities in color, gray level, texture etc. canny, sobel are edge detection operator which are basically used for detecting an edge [10] [12] [14] [15].

The simplest method of image segmentation is called the thresholding method. By using threshold value, image binarization is performed. Threshold is used to separate the region in an image with respect to the object, which is to be analysed and this is based on the variation of intensity between the object pixel and background pixel.

Another approach is a region growing method used for segmentation .In the present research work after enhancement of image, the region of each rice grain in an image is detected using region growing. More specifically, image segmentation is the process of assigning a label to every pixel in an image such that pixels with the same label share certain visual characteristics.

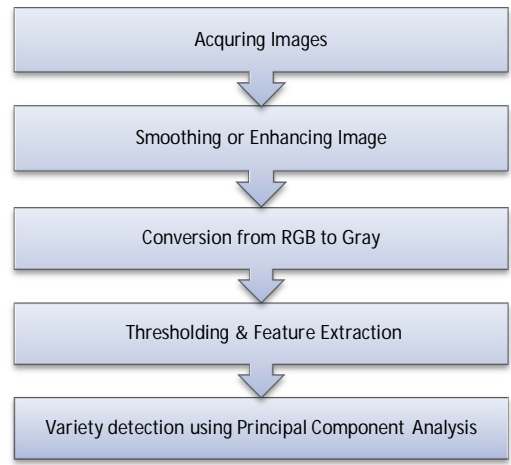

Fig. 2  Flow chart for general steps

V. FEATURE EXTRACTION

A. *Morphological Feature Extraction.*

Extraction of quantitative information from image is deal with feature extraction. The work of categorization and recognizition of object based on various feature such as morphological feature, color feature extraction and textural feature [16] [17] [18] [19] . In the present research work morphological feature are extracted.

Algorithms were developed in windows environment using MATLAB 7.7.0 programming language to extract morphological features of individual basmati rice grains. The following morphological features were extracted from images of individual basmati rice grains:

**Area**: The algorithm calculates the actual number of pixels inside and including the seed boundary (mm2/pixel).

**Major Axis Length:** It was the distance between the end points of the longest line that could be drawn through the seed. The major axis endpoints were found by computing the pixel distance between every combination of border pixels in the seed boundary.

**Minor Axis Length:** It was the distance between the endpoints of the longest line that could be drawn through the seed while maintaining perpendicularity with the major axis.

**Eccentricity:** specifies the eccentricity of the ellipse that has the same second-moments as the region. The eccentricity is the ratio of the distance between the foci of the ellipse and its major axis length. The value is between 0 and 1.

**Perimeter:** It was the total pixel that constitutes the edge of object. It helps in locate the object & provide information about the shape of the object i.e. counting the number of '1' pixel that have '0' pixel of neighbour.

## VI. RESULT AND DISCUSSION

A. *Image Sample*

Sample image of classic basmati rice and feature extracted from grains is shown in Table 1.

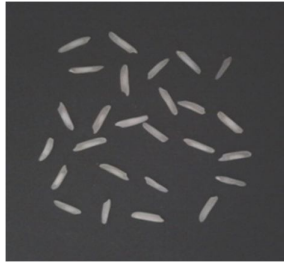

Fig. 3 Sample of Classic Basmati Rice

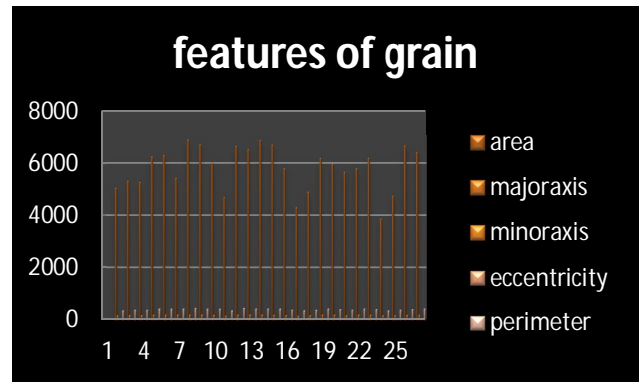

Fig. 4 Different feature extracted from sample of Classic Basmati Rice

B. *Sample of final data of Classic, Rozana and Mini basmati rice grain.*

TABLE I
Feature Table

| Sno. | Area | Major axis | Minor axis | Eccentricity | Perimeter |
|---|---|---|---|---|---|
| 1 | 5043 | 156.6652 | 42.9724 | 0.9616 | 357.262 |
| 2 | 5314 | 163.9978 | 42.5703 | 0.9657 | 372.132 |
| 3 | 5279 | 165.1427 | 41.3308 | 0.9682 | 371.4041 |
| 4 | 6251 | 178.3676 | 46.1127 | 0.966 | 410.3747 |
| 5 | 6310 | 194.7024 | 41.7907 | 0.9767 | 429.6884 |
| 6 | 5424 | 191.6559 | 36.9983 | 0.9812 | 412.5513 |
| 7 | 6899 | 197.3961 | 45.8897 | 0.9726 | 441.1615 |
| 8 | 6714 | 190.3464 | 46.0767 | 0.9703 | 429.5879 |
| 9 | 6009 | 180.7577 | 43.8813 | 0.9701 | 412.1737 |
| 10 | 4696 | 151.3532 | 40.5578 | 0.9634 | 342.3087 |
| 11 | 6648 | 195.8673 | 44.3653 | 0.974 | 431.7128 |
| 12 | 6535 | 186.8246 | 45.6589 | 0.9697 | 424.1909 |
| 13 | 6873 | 188.6419 | 48.2112 | 0.9668 | 423.9655 |
| 14 | 6688 | 182.1961 | 47.6576 | 0.9652 | 409.0782 |
| 15 | 5803 | 176.4117 | 42.9413 | 0.9699 | 383.0854 |
| 16 | 4303 | 152.9522 | 37.1873 | 0.97 | 340.9605 |
| 17 | 4898 | 164.459 | 38.9026 | 0.9716 | 362.7006 |
| 18 | 6190 | 191.0569 | 42.052 | 0.9755 | 421.2447 |
| 19 | 5957 | 187.2706 | 41.4544 | 0.9752 | 403.4701 |
| 20 | 5663 | 162.2814 | 46.1867 | 0.9586 | 376.6762 |
| 21 | 5796 | 188.0324 | 39.6944 | 0.9775 | 412.3747 |
| 22 | 6179 | 181.468 | 44.7413 | 0.9691 | 404.1737 |
| 23 | 3872 | 156.0474 | 32.6108 | 0.9779 | 346.4752 |
| 24 | 4751 | 174.3225 | 35.1834 | 0.9794 | 378.4924 |
| 25 | 6679 | 185.0608 | 46.9024 | 0.9674 | 399.4701 |
| 26 | 6425 | 183.8406 | 45.8627 | 0.9684 | 413.5219 |

TABLE II
Sample Table

| Basmati Variety | SNo | No. of grains | Average | Time(Sec) |
|---|---|---|---|---|
| Classic(100% basmati) | S1 | 49 | 2.1635e-013 | 1.956056 |
| | S2 | 29 | 3.6066e-013 | 2.013604 |
| | S3 | 72 | -1.2632e-014 | 2.086206 |
| Rozana | S1 | 23 | -1.2852e-013 | 1.974158 |
| | S2 | 37 | 1.2290e-014 | 2.571046 |
| | S3 | 67 | -1.6968e-014 | 2.029567 |
| Mini(50% basmati) | S1 | 16 | -1.5987e-014 | 1.937171 |
| | S2 | 60 | 4.5475e-014 | 1.976318 |
| | S3 | 124 | -1.4509e-013 | 1.938460 |

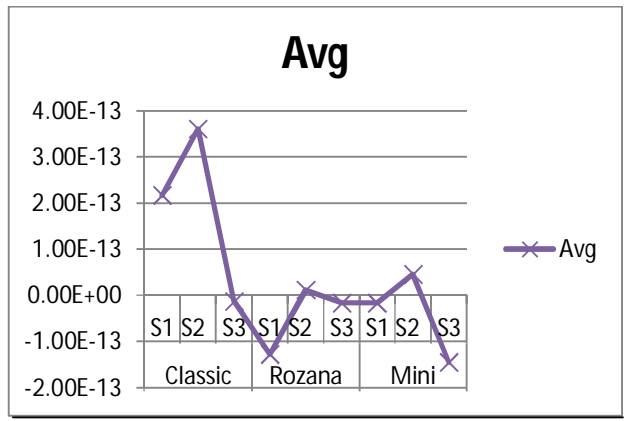

Fig. 6 Showing variety of Basmati rice Sample with their average.

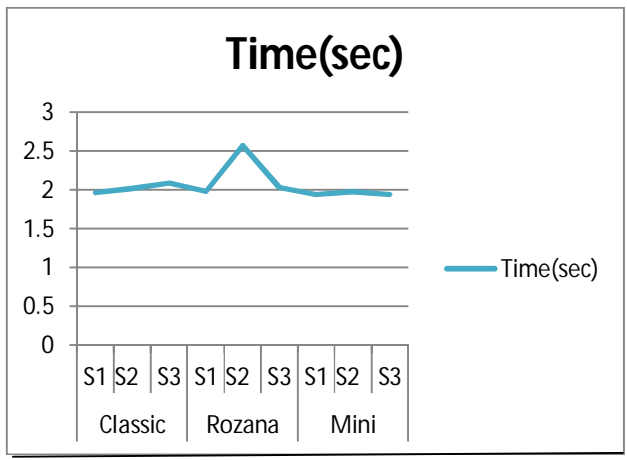

Fig. 7 Showing time taken by each variety of Basmati rice Sample.

TABLE III
Test Table

| Basmati variety | Sample no | No. of grain | output | | |
|---|---|---|---|---|---|
| | | | Classic | Rozana | mini |
| classic | CTs1 | 29 | √ | | |
| classic | CTs2 | 82 | | | √ |
| classic | CTs3 | 49 | √ | | |
| classic | CTs4 | 28 | √ | | |
| Classic | CTs5 | 72 | √ | | |
| rozana | RTs1 | 22 | | √ | |
| rozana | RTs2 | 37 | | √ | |
| rozana | RTs2 | 67 | | | √ |
| rozana | RTs4 | 61 | | √ | |
| Mini | MTs1 | 16 | | | √ |
| Mini | MTs2 | 62 | | | √ |
| Mini | MTs3 | 68 | | | √ |
| Mini | MTs4 | 124 | | √ | |
| Mini | MTs5 | 69 | | | √ |

TABLE IV
*Accuracy Table*

| Variety | Total no of sample | Accuracy |
|---|---|---|
| Classic | 5 | 80% |
| Rozana | 4 | 75% |
| Mini | 5 | 80% |
| Overall | 14 | 79% |

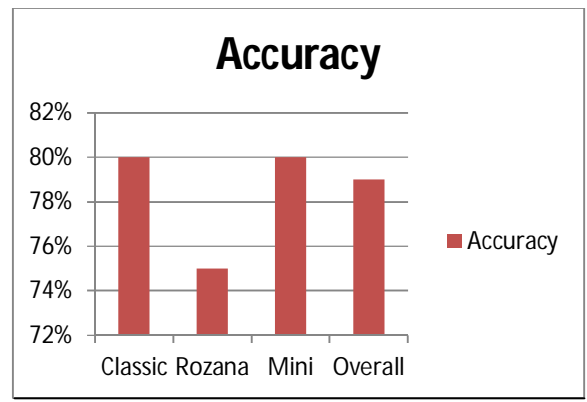

Fig. 8 Showing Overall and individual Accuracy achieved for classification of variety of Basmati rice Sample.

According to Table II, it is observed that the value of Principal component is depending upon the number of rice grains present in sample. For classification we use best of two average values to calculate final value. And for classification K-NN Classifier is used.

Table III and IV shows the output and the effectiveness of proposed method.

## V. CONCLUSIONS

The Experimental result shows that the proposed algorithm work effectively for samples containing less number of rice grain for all variety of basmati rice. The accuracy of proposed algorithm for classic 80%, rozana 75% and mini 80%, and the overall accuracy of basmati rice is 79%.

ACKNOWLEDGMENT

We would like to express our special thanks to Mr. V C Jain for the sample of basmati rice variety as per the requirement desired by us and also thankful for their helpful advice.

REFERENCES

[1] Chetan V.Maheshwari, "Machine Vision Technology for Oryza Sativa L.(RICE)," IJAREEIE, vol. 2, no. 7, pp. 2893-2900, july 2013.

[2] Chetan V.Maheshwari, "Quality Assesment of ORYZA SATIVA SSP INDICA (Rice)using computer vision," IJIRCCE, vol. 1, no. 4, pp. 1107-1115, june 2013.

[3] D.Swain ,Sanjukta Das,S.G.Sharma & O.N.Singh H.N.Subudhi, "Studies on Grain Yield,Physico-Chemical and Cooking Characters of Elite Rice Varieties(Oryza Sativa.L)in Eastern India," Journal of Agriculture Science, vol. 4, no. 12, pp. 269-275, 2012.

[4] Sachin Patel Bhavesh B.Prajapati, "Classification of Indian Basmati Rice using Digital Image Processing As per Indian Export Rules," International Research Jornal of Computer Science Engineering and Application, vol. 2, no. 1, pp. 234-237, january 2013.

[5] Sachin Patel Bhavesh B.Prajapati, "Algorithmic approach to quality analysis of Indian Basmati Rice using Digital Image Processing," IJETAE (ISSN 2250-2459,ISO 90001:2008certified journal), vol. 3, no. 3, pp. 503-504, March 2013.

[6] Yaduvir singh and sweta tripathi Pramod kumar pandey, "Image Processing using Principal Component Analysis," International Journal Of Computer Application (0975-8887), vol. 15, no. 4, pp. 37-40, february 2011.

[7] Abhishek Banerjee, "Impact of Principal component analysis in the application of image processing," IJARCSSE(ISSN: 2277 128X), vol. 2, no. 1, January 2012.

[8] V.K.Banga Jagdeep singh Aulakh, "Percentage Purity of Rice Samples by Image Processing," in International conference on Trends in Electrical ,Electronics and Power Engineering(ICTEEP 2012), Singapore, july 15-16,2012, pp. 102-104.

[9] Mrs.Uzma Ansari sanjivani shantaiya, "Identification of Food grains and its quality using pattern clasification," vol. 2 , no. 2,3,4, pp. 70-73, december 2010.

[10] Abirami.S,Vishnu Priya.K,Rubalya Valantina.S. Neelamegam.P, "Analysis Of rice granules using Image Processing and Neural Network," in IEEE conference on Information and Communication Technologies(ICT2013), 2013, pp. 879-884.

[11] M.Shahid Naweed ,M NAdim Asif and S.Irfan Hyder Imran S.Bajwa, "Feature based Image Classification by using Principal Component Analysis," CIST-journal of Graphica,vision and image processing.

[12] Kavindra R.jain,Chintan K.Modi Chetan v.Maheshwari, "Non Destructive Quality Analysis of India Gujrat-17 ORYZA SATIVA SSP INDICA(Rice)using Image Processing," IJCES ISSN:22550:3439, vol. 2, no. 3, pp. 48-54, march 2012.

[13] B.Karthikeyan R.Yogamangalam, "Segmentation Technique Comparision in Image Processing," IJET(ISSN :0975-4024), vol. 5, no. 1, pp. 307-313, feb-march 2013.

[14] Gurpreet Singh,Amandeep Kaur Pooja Sharma, "Different Technique Of Edge Detection In Digital Image Processing," IJERA(ISSN:2248-9622), vol. 3, no. 3, pp. 458-461, May-Jun 2013.

[15] G.Kishorebabu,K.sujatha M.Kalpana, "Extraction of Edge Detection using Image Processing Technique," IJCER, vol. 2, no. 5, pp. 1562-1566, september 2012.

[16] Dr.M.Siddappa Harish S Gujjar, "Amethod for identification of rice grainn of india and its quality using pattern classification," IJERA ISSN:2248-9622, vol. 3, no. 1, pp. 268-273, january-february 2013.

[17] Dayanand SAVAKAR, "Recognition and Classification of similar looking food grain images using Artificial Neural Networks," Journal of Applied computer Science & Mathematics , no. 13, pp. 61-65, June 2012.

[18] Chathurika sewwandi silva and upul sonnadara, "Classifiction Of Rice Grain Using Neural Network," in Proceeding of Technical session , SriLanka, 2013, pp. 9-14.

[19] R.M.Carter,Y.yan D.M.Hobson, "Characterisation and Identification of Rice Grains through Digital Image Analysis," in Technology Conference-IMTC, warsaw,Poland, 2007, pp. 1-5.

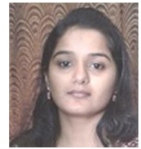

**Rubi Kambo** is a M.Tech Scholar of RCET, Bhilai (C.G), India. She did his M.C.A (Master of Computer Application) from Punjab Technical University at Jalandhar (Punjab), India.

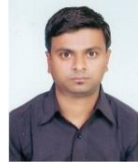

**Mr. Amit Yerpude is** working as Associate Professor in Department Of computer Science & Engg. , RCET, Bhilai (Chhattisgarh), India. He has received M.Tech in Computer science from CSVTU University, Bhilai (C.G).He has published much research paper in international journals and presented several research   papers in international conferences.